\documentclass{article}

\usepackage[preprint]{neurips_2022}

\usepackage[utf8]{inputenc} 
\usepackage[T1]{fontenc}    
\usepackage{hyperref}       
\usepackage{url}            
\usepackage{booktabs}       
\usepackage{amsfonts}       
\usepackage{nicefrac}       
\usepackage{microtype}      
\usepackage{xcolor}         
\usepackage{graphicx}
\graphicspath{ {./images/} }

\title{Applied Deep Learning to Identify and Localize Polyps from Endoscopic Images }

\author{%
Sumedh Vilas Datar ,
Chandana Raju,
Kushala Hari,
Kavin Vijay,
Suma Ningappa
}

\begin{document}
\maketitle

\begin{abstract}
Deep learning based neural networks have gained popularity for a variety of biomedical imaging applications. In the last few years several works have shown the use of these methods for colon cancer detection and the early results have been promising. These methods can potentially be utilized to assist doctor's and may help in identifying the number of lesions or abnormalities in a diagnosis session. From our literature survey we found out that there is a lack of publicly available labeled data. Thus, as part of this work, we have aimed at open sourcing a dataset which contains annotations of polyps and ulcers. This is the first dataset that's coming from India containing polyp and ulcer images. The dataset can be used for detection and classification tasks. We also evaluated our dataset with several popular deep learning object detection models that's trained on large publicly available datasets and found out empirically that the model trained on one dataset works well on our dataset that has data being captured in a different acquisition device.       
\end{abstract}

\section{INTRODUCTION}
Colorectal cancer is the third most diagnosed cancer worldwide and also the third most common cancer diagnosed in the United States  in both men and women [1]. Polyps are the initial stages of colorectal cancer, which is a growth on the inner lining of the colon. Overtime, some kinds of polyps can change into cancer. Currently doctors use an endoscopic camera to see the polyps from a camera feed and take snapshots during diagnosis. After diagnosis, they revisit  all the captured images and count the number of polyps. This process is very tiresome since they have to go through every image and we found a need to solve this problem with computer aided assistance.

Deep Learning has emerged as a potent tool to look for accurate detection of polyp from endoscopic images. There are several works related to identification of polyps. A lot of research has been done in the space of classification and segmentation[4-8,13]. 

In this paper, we present an annotation dataset for polyp segmentation and this is the first dataset from India. We further use a pre-trained model to train the model on the popular KVASIR-SEG[11] data and apply and show the efficacy on our dataset. From our evaluation we conclude that model trained on KVASIR-SEG dataset translate well on our dataset



\section{DATASET}

The Kvasir-SEG[11] contains a subset of the original Kvasir dataset[8]. The Kvasir-SEG dataset contains the annotations for all the polyps. The dataset also has masks. There are other publicly available datasets as explained in [11,13]. But in every dataset the number of polyp images are very less. 

Our dataset contains images containing both polyps and ulcers. The Endoscopic images for this work have been obtained from Department of Gastroenterology, Kempegowda Institute of Medical Sciences (KIMS), Bangalore, India. The department uses Olympus Exera CV-180 for the acquisition of the images. We received 20,000 images in total and out of 20,000 images we found out 180 raw images of polyps and 232 raw images of ulcers and the data is hand annotated and verified with the field expert. This shows the amount of effort it takes to find polyp images. We used VOTT[12] annotation tool to annotate the data.

A dataset from India is not publicly available and we are the first to open source this dataset and provide access to researchers to benchmark their results on varied datasets and add additional collection of polyp data to the existing publicly available datasets.

The data is open sourced and here s the link to access the data
https://github.com/sumedhvdatar/deep-endoscopy

\section{EXPERIMENTAL SETUP and EVALUATION}
We used Faster RCNN[14] object detection model with Detectron2[15] Pytorch Framework. We trained the model on google colab notebook and the GPU has 15gb RAM. We trained our model on different popular object detection models as shown in Table-1 and having a learning rate of 0.0025. The train data had 1000 images and test data had 180 images at the time of running the experiment. We used Intersection Over Union (IOU) as the evaluation metric to evaluate the performance of the object detection model.

\begin{center}
\begin{tabular}{||c c c ||} 
 \hline
 Model Name & Training Loss & IOU on KIMS Dataset \\ [0.5ex] 
 \hline\hline
 Faster-rcnn-R-50-C4-1x & 0.502 & 0.56 \\ 
 \hline
 Faster-rcnn-R-50-DC5-1x & 0.4 & 0.6\\
 \hline
 Faster-rcnn-R-50-FPN-3x & 0.284 & 0.61\\
 \hline
 Faster-rcnn-R-101-FPN-3x & 0.29 & 0.63 \\
 \hline
\end{tabular}
\end{center}
Table 1 - Benchmark Test IOU on different object detection models

\begin{center}
\includegraphics[width=3cm, height=3cm]{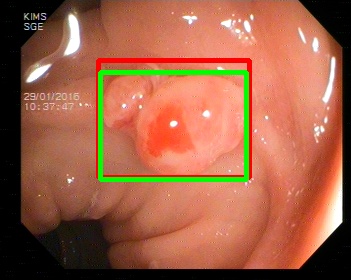}
\includegraphics[width=3cm, height=3cm]{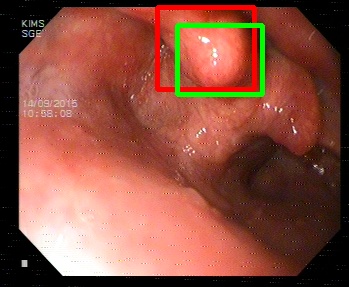}
\includegraphics[width=3cm, height=3cm]{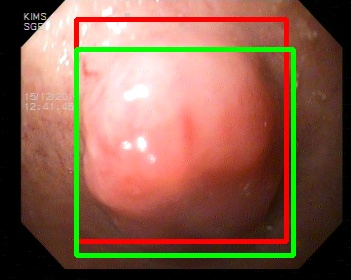}

Figure 1- Shows the results from ground truth and predicted boxes. Green boxes are human labelled boxes and red boxes are predicted by the model
\end{center}

\begin{center}
\includegraphics[width=3cm, height=3cm]{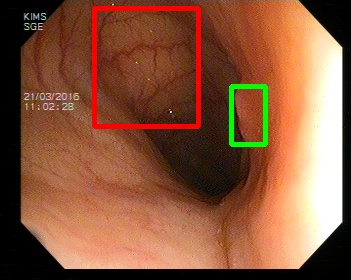}

Figure 2- Shows the result where the model has detected at a wrong location
\end{center}

\section{RESULTS AND DISCUSSIONS}

We evaluated our dataset containing polyp images with several FasterRCNN models after training to convergence as shown in Table -1. We see that the highest Intersection Over Union that we received was 0.63 compared to the benchmark from [11] having 0.77. In Figure 1 we can see how the model has localized the polyps in the image. In Figure 2 we can see where the model missed in identification of polyps. Over all for every model the training time took less than 15mins.

\section{IMPACT TO THE SOCIETY}

Applied deep learning in identification of abnormalities from internal organs has become very vital. Computer aided diagnosis helps in better diagnosis and better healthcare. Our work showcases a technique to train models on publicly available data and deploy it onto a system that helps doctors in assisting and making better diagnosis of patient. 

The same concept can be applied to different use cases like identification of abnormalities in different parts of the body thereby starting of by training a model on publicly available data and very little custom data to evaluate the performance of the model.  

Overall this will help in making sure the abnormalities are discovered earlier and thus patients can be suggested precautionary measures earlier and provide better healthcare and treatment. However the model might miss a particular polyp as explained in Figure-2 and that the prediction capability of the model improves with more number of data and this is just to support the doctor and the doctors decision is the final decision. It only assists doctors in better diagnosis. 

\section{ACKNOWLEDGEMENT}
We would like to thank Dr Sahadev from The Department of Gaestroentology KIMS Hospital , Bangalore  , India for giving us an opportunity to understand the diagnosis process.

\section{CONCLUSION}

In this work we present a new open source dataset that is human annotated and labelled that can be used for detection and classifications tasks. We also explain the current pain point that doctors face and show how we solved the problem. We also demonstrate that the model trained on one dataset with one acquisition device can be used to predict polyps from images with our dataset having a different acquisition device.

\section*{REFERENCES}

{
\small
[1]    “Colorectal Cancer - Statistics,” Cancer.Net, Jun. 25, 2012. https://www.cancer.net/cancer-types/colorectal-cancer/statistics (accessed Sep. 05, 2022) \\

[2]    “What Is Colorectal Cancer?” https://www.cancer.org/cancer/colon-rectal-cancer/about/what-is-colorectal-cancer.html (accessed Sep. 05, 2022).\\

[3]    “Colorectal Cancer Statistics.” https://www.cancer.org/cancer/colon-rectal-cancer/about/key-statistics.html (accessed Sep. 05, 2022).\\

[4]    C. Y. Eu, T. B. Tang, C.-H. Lin, L. H. Lee, and C.-K. Lu, “Automatic Polyp Segmentation in Colonoscopy Images Using a Modified Deep Convolutional Encoder-Decoder Architecture,” Sensors , vol. 21, no. 16, Aug. 2021, doi: 10.3390/s21165630.\\

[5] Imagenet classification with deep convolutional neural networks Alex Krizhevsky, Ilya Sutskever, Geoffrey E Hinton \\

[6]C.-M. Hsu, C.-C. Hsu, Z.-M. Hsu, F.-Y. Shih, M.-L. Chang, and T.-H. Chen, “Colorectal Polyp Image Detection and Classification through Grayscale Images and Deep Learning,” Sensors , vol. 21, no. 18, Sep. 2021, doi: 10.3390/s21185995.\\

[7]Ozawa T, Ishihara S, Fujishiro M, Kumagai Y, Shichijo S, Tada T. Automated endoscopic detection and classification of colorectal polyps using convolutional neural networks. Therap Adv Gastroenterol. 2020 \\

[8]“Colonic Polyp Detection in Endoscopic Videos With Single Shot Detection Based Deep Convolutional Neural Network.” https://ieeexplore.ieee.org/document/8731913 (accessed Sep. 05, 2022).\\

[9]D. Shen, G. Wu, and H.-I. Suk, “Deep Learning in Medical Image Analysis,” Annu. Rev. Biomed. Eng., vol. 19, p. 221, Jun. 2017.\\

[10]KVASIR: A Multi-Class Image Dataset for Computer Aided Gastrointestinal  Konstantin Pogorelov , Kristin Ranheim Randel, Carsten Griwodz , Sigrun Losada Eskeland, Thomas de Lange, Dag Johansen, Concetto Spampinato, Duc-Tien Dang-Nguyen, Mathias Lux, Peter Thelin Schmidt, Michael Riegler, Pal Halvorsen

[11] Kvasir-SEG: A Segmented Polyp Dataset
Debesh Jha, Pia H. Smedsrud, Michael A. Riegler, P al Halvorsen,
Thomas de Lange, Dag Johansen, and Havard D. Johansen \\

[12]https://github.com/microsoft/VoTT\\

[13]Jha, Debesh et al. “Real-Time Polyp Detection, Localization and Segmentation in Colonoscopy Using Deep Learning.” 
IEEE access : practical innovations, open solutions Mar. 2021 \\

[14] Faster R-CNN: Towards real-time object detection with region proposal networks Shaoqing Ren, Kaiming He, Ross Girshick, Jian Sun 

[15]@misc{wu2019detectron2,
  author =       {Yuxin Wu and Alexander Kirillov and Francisco Massa and
                  Wan-Yen Lo and Ross Girshick},
  title =        {Detectron2},
  howpublished = {\url{https://github.com/facebookresearch/detectron2}},
  year =         {2019}
}

}


Optionally include extra information (complete proofs, additional experiments and plots) in the appendix.
This section will often be part of the supplemental material.

\end{document}